# New Worst-Case Upper Bound for #2-SAT and #3-SAT with the Number of Clauses as the Parameter

**Junping Zhou[1,2], Minghao Yin[2,3*], Chunguang Zhou[1]**

[1]College of Computer Science and Technology, Jilin University, Changchun, P. R. China, 130012
[2]College of Computer, Northeast Normal University, Changchun, P. R. China, 130117
[3]Key Laboratory of Symbolic Computation and Knowledge Engineering of Ministry of Education, Changchun, P. R. China 130012
zhoujp08@mails.jlu.edu.cn, mhyin@nenu.edu.cn, cgzhou@jlu.edu.cn

**Abstract**

The rigorous theoretical analyses of algorithms for #SAT have been proposed in the literature. As we know, previous algorithms for solving #SAT have been analyzed only regarding the number of variables as the parameter. However, the time complexity for solving #SAT instances depends not only on the number of variables, but also on the number of clauses. Therefore, it is significant to exploit the time complexity from the other point of view, i.e. the number of clauses. In this paper, we present algorithms for solving #2-SAT and #3-SAT with rigorous complexity analyses using the number of clauses as the parameter. By analyzing the algorithms, we obtain the new worst-case upper bounds $O(1.1892^m)$ for #2-SAT and $O(1.4142^m)$ for #3-SAT, where $m$ is the number of clauses.

## Introduction

Propositional model counting or #SAT is the problem of computing the number of models for a given propositional formula, i.e., the number of distinct truth assignments to variables for which the formula evaluates to *true*. Nowadays, efficient model counting algorithms have opened up a range of applications. For example, various probabilistic inference problems can be translated into model counting problems (cf. Park 2002; Sang *et al*. 2005). #SAT problem can be viewed as a generalization of the well-known canonical NP-complete problem of Propositional Satisfiability (SAT), which has been well studied. Actually, model counting has been proved to be #P-complete, harder than standard SAT problems (Bacchus *et al.* 2003). Therefore, improvements in exponential time bounds are crucial in determining the size of model counting problem that can be solved. Even a slight improvement from $O(c^k)$ to $O((c-\varepsilon)^k)$ may significantly increase the size of the problem being tractable.

Recently, tremendous efforts have been made on efficient #SAT algorithms with complexity analyses. By introducing independent clauses and combining formulas, Dubois (1991) presented a #SAT algorithm which ran in $O(1.6180^n)$ for #2-SAT and $O(1.8393^n)$ for #3-SAT, where $n$ is the number of variables of a formula. Based on a more elaborate analysis of the relationship among the variables, Dahllof et al. (2002) presented algorithms running in $O(1.3247^n)$ for #2-SAT and $O(1.6894^n)$ for #3-SAT. Furer et al. (2007) presented an algorithm performing in $O(1.246^n)$ for #2-SAT by using a standard reduction. Further improved algorithms in (Kutzkov 2007) presented a new upper time bound for the #3-SAT ($O(1.6423^n)$), which is the best upper bound so far.

Different from complexity analyses regarding the number of variables as the parameter, Hirsch (2000) introduced a SAT algorithm with a time bound $O(1.239^m)$, where $m$ is the number of clauses of a formula. An improved algorithm for SAT with an upper bound $O(1.234^m)$ was proposed in (Masaki 2005). Skjernaa (2004) presented an algorithm for Exact Satisfiability with a time bound $O(2^m)$. Bolette (2006) addressed an algorithm for Exact Satisfiability with a time bound $O(m!)$.

Similar to the SAT problem, the time complexity of #SAT problem is calculated based on the size of the #SAT instances, which depends not only on the number of variables, but also on the number of clauses. Therefore, it is significant to exploit the time complexity from the other point of view, i.e. the number of clauses. However, so far all algorithms for solving #SAT have been analyzed based on the number of variables. And to our best knowledge, it is still an open problem that analyzes the #SAT algorithm with the number of clauses as the parameter.

The aim of this paper is to exploit new upper bounds for #2-SAT and #3-SAT using the number of clauses as the parameter. We provide algorithms for solving #2-SAT and #3-SAT respectively. The algorithm for #2-SAT employs a new principle, i.e. the five-vertex principle, to simplify formulae. This allows us to eliminate variables whose

---

degree is 3, and therefore improves the efficiency of the algorithm. In addition, by transforming a formula into a constraint graph, we propose some detailed analyses between the adjacent variables in the constraint graph, which provides us theoretical foundations for choosing better variables to branch. By analyzing the algorithm, we obtain the worst-case upper bound $O(1.1892^m)$ for #2-SAT. For the algorithm solving #3-SAT, we adopt a different strategy, first simplifying the 3-clause formulae into 2-clause formulae, then solving these 2-clasue formulae by the algorithm for #2-SAT. To demonstrate that this strategy is efficient, we give a deep analysis and obtain the worst-case upper bound $O(1.4142^m)$ for #3-SAT.

## Problem Definitions

We now describe the definitions that will be used in this paper. A literal is either a Boolean variable $x$ or its negation $\neg x$. If a literal is $l$, the negation of the literal is $\neg l$. A clause is a disjunction of literals which doesn't contain a complementary pair $x$ and $\neg x$ simultaneously. A 2-clause is the clause that contains exactly two literals. A 3-clause is the corresponding clause. The length of a clause $C$ is the number of literals in it. A clause $C$ is a unit clause if the length of the clause is 1. We call the literal in the unit clause is the unit literal. A $k$-SAT formula $F$ in Conjunction Normal Form (CNF) is a conjunction of clauses, each of which contains exactly $k$ literals. Any Boolean variable $x_i$ in $F$ can take a value *true* or *false*. A truth assignment for $F$ is a map that assigns each variable a value. The satisfying assignment, called model, is the truth assignment that makes $F$ evaluated to *true*. The propositional model counting or #SAT problem is to determine the number of satisfying assignments for a formula. #2-SAT is the problem of computing the number of satisfying assignments of a 2-SAT formula and #3-SAT is the corresponding problem for a 3-SAT formula.

A formula $F$ in CNF can be expressed as an undirected graph called constraint graph. In the constraint graph $G$, the vertexes are the variables of $F$ and the edges between two vertexes if the corresponding variables appear together in some clause of $F$. The degree of a vertex is the number of edges incident to the vertex. The degree of a Boolean variable $x$, represented by $\varphi(x)$, is the degree of the corresponding vertex. The degree of a formula $F$, denoted by $\varphi(F)$, is the maximum degree of variables in $F$. We say a formula is a cycle or a path whenever the constraint graph forms a cycle or a path. We define M($F$) as the number of models of the formula $F$, $m$ as the number of clauses in $F$, and $n$ as the number of variables $F$ contains.

After specifying the definitions, we present some basic rules for solving #SAT problem. Suppose constraint graph $G$ can be partitioned into disjoint components $G_1$ and $G_2$ where there is no edge connecting a vertex in $G_1$ to a vertex in $G_2$, i.e. the variables corresponding to vertexes in $G_1$ and $G_2$ are mutually disjoint. Let $F_1$ and $F_2$ be the sub-formulae of $F$ corresponding to the two components $G_1$ and $G_2$. Then,

$$M(F)=M(F_1)\times M(F_2) \qquad (1)$$

Given a formula $F$, the basic strategy of Davis-Putnam-Logemann-Loveland (DPLL) computing the models of $F$ is to arbitrarily choose a variable $x$ that appears in $F$. Then,

$$M(F)=M(F\wedge x)+M(F\wedge \neg x) \qquad (2)$$

In order to determine the number of models of $F \wedge l$, we adopt the unit literal rule which assigns the unit literal *true*. The result of the unit literal rule, denoted by $F|_l$, can be obtained by (1) removing all clauses containing the literal $l$ from $F$, and (2) deleting all occurrences of $\neg l$ from the other clauses. However, when we apply the unit literal rule to the two sub-formulae $F \wedge x$ and $F \wedge \neg x$, variables that appear in $F$ may not appear in the simplified version $F|_x$ and $F|_{\neg x}$, which may make M($F \wedge x$) and M($F \wedge \neg x$) wrong. For example, let $F=x \vee y \vee z$ and we choose $x$ to branch, then

$$M(F)=M(F\wedge x)+M(F\wedge \neg x) \qquad (3)$$

When the sub-formula $F \wedge x$ is simplified by the unit literal rule, variable $y$ and $z$ are eliminated. Thus, M($F|_x$)=1. When the same rule is applied to $F$ and $F \wedge \neg x$, we obtain M($F$)=7 and M($F|_{\neg x}$)=3. It is obvious that M($F$) ≠ M($F \wedge x$)+M($F \wedge \neg x$). Therefore, we introduce a variable set $R$ to record the eliminated variables, which we will give a detailed description in the next section.

## The Complexity Measure

In this subsection, we explain how we compute the complexity of our algorithms. At first, we give a notion called branching tree. The branching tree (Hirsch 2000) is a hierarchical tree structure with a set of nodes, each of which is labeled with a formula. Suppose a node is labeled with a formula $F$, then its sons are labeled with the sub-formulae $F_1, F_2, \ldots, F_j$, each of which is obtained by assigning a value to one of variables in $F$. From the definition we can see that the process of constructing a branching tree is the same as the process of executing DPLL-style algorithms, therefore, we use the branching tree to estimate the time complexity.

In the branching tree, every node has a branching vector. Let us consider a node labeled with $F_0$ and its children nodes labeled with $F_1, F_2, \ldots, F_k$. The branching vector of the node labeled with $F_0$ is $(r_1, r_2,\ldots, r_k)$, where $r_i=f(F_0)-f(F_i)$ ($f(F_0)$ is the number of clauses of $F_0$). The value of the branching vector of a node, called branching number ($\lambda(r_1, r_2,\ldots, r_k)$), is obtained from the positive root of the following equation.

$$\sum_{i=1}^{k} x^{-r_i} = 1 \qquad (4)$$

We define the maximum branching number of nodes in the branching tree as the branching number of the branching tree, expressed by max $\lambda(r_1, r_2,\ldots, r_k)$. The branching number of a branching tree has an important relationship

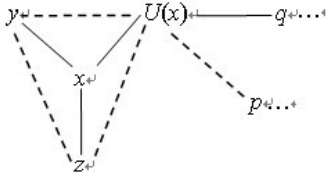

Figure1: A constraint graph where $L_F(x)=1$; the solid lines indicate the end point variables of each solid line appear together in some clause of $F$; the dashed lines indicate the end points variables of each dashed line may appear together in some clause of $F$.

with the running time ($T(m)$) of DPLL-style algorithms. At first, we assume that the running time of DPLL-style algorithms performing on each node is in polynomial time. Then we obtain the following inequality.

$$T(m) \leq (\max \lambda (r_1, r_2, \ldots, r_k))^m \times \text{poly}(F)$$
$$= (\max \sum_{i=1}^{k} T(m-r_i))^m \times \text{poly}(F) \quad (5)$$

where $m$ is the number of clauses in the formula $F$, $\text{ploy}(F)$ is the polynomial time executing on the node $F$, and

$$\lambda (r_1, r_2, \ldots, r_k) = \sum_{i=1}^{k} T(m-r_i) \quad (6)$$

In addition, if a #SAT problem recursively solved by the DPLL-style algorithms, the time required doesn't increase, for

$$\sum_{i=1}^{k} T(m_i) \leq T(m) \text{ where } m = \sum_{i=1}^{k} m_i \quad (7)$$

where $m$ is the number of clauses, $m_i$ is the number of clauses in the sub-formula $F_i$ ($1 \leq i \leq k$) of the formula $F$. Note that when analyzing the running time of our algorithms, we ignore the polynomial factor so that we assume that all polynomial time computations take $O(1)$ time in this paper.

## Algorithm for #2-SAT

In this section, we present the algorithm $MC_2$ for #2-SAT and prove an upper bound $O(1.1892^m)$. Firstly we address some preliminaries used in this part.

### Preliminaries

We begin the subsection by specifying some notions similar to that proposed in (Dahllöf et al. 2002). $\{x_1, x_2\}$ represents a formula which is composed of the variables $x_1$ and $x_2$. Given a formula $F$ expressed as a constraint graph $G$ and a vertex $x$, $L_F(x)$ is the number of vertexes which are not only adjacent to $x$ but also adjacent to other vertexes not in the neighborhood of $x$, i.e.,

$$L_F(x) = |\{u | (u,v) \in G \wedge u \in N_G(x) \wedge v \notin N_G(x) \cup \{x\}\}| \quad (8)$$

where $u$ and $v$ are vertexes, $(u, v)$ is an edge, and $N_G(x)$ is the neighborhood of $x$ in the constraint graph $G$. When $L_F(x)=1$, the unique variable corresponding to the vertex is denoted by $U(x)$, just as Figure1 describes.

### Helpful Function and Principle

The subsection discusses some functions and principles used for simplifying the formulae. The first function $unit(F, l)$ in Figure 2 is to record the variables which appear in the unit clauses after assigning the literal $l$ $true$. $var(l)$ denotes the variable forming the literal $l$. The second function $\Omega (F, R, l)$ in Figure 3 recursively executes the unit literal rule. The function takes as input the formula $F$, a variable set $R$ recording the eliminated variables, and a literal $l$ being assigned $true$. The detailed process of the function is presented as follows. (1) Remove all clauses containing literal $l$ from $F$; (2) Delete all occurrences of the negation of literal $l$ from the other clauses; (3) Perform the process as far as possible. Finally, the function returns a simplified formula and a new set $R$.

---

Function $unit(F, l)$
1. If $F$ is empty, return $V = \varnothing$.
2. If there exists a clause $\neg l \vee l'$, add $var(l')$ to $V$.
3. Do $unit(F, l)$ until it never adds variables into $V$.
4. Return $V$.

---

Figure 2: Function $unit$

---

Function $\Omega (F, R, l)$
1. If there exists a clause $l \vee l_1 \vee l_2 \vee \ldots \vee l_n$ in $F$, remove $l \vee l_1 \vee l_2 \vee \ldots \vee l_n$ from $F$.
2. If there exists a clause $\neg l \vee l'_1 \vee l'_2 \vee \ldots \vee l'_k$ in $F$, remove $\neg l$ from $\neg l \vee l'_1 \vee l'_2 \vee \ldots \vee l'_k$.
3. Update the variable set $R$ and the formula $F$.
4. Do $\Omega (F, R, l)$ until $F$ doesn't contain $l$ and $\neg l$.
5. Return $F$ and $R$.

---

Figure 3: Function $\Omega$

Now we concentrate on the introduction of the five-vertex principle whose applicable condition is described in Figure 4. Supposing in a 2-SAT formula $F$, one of the maximum degree variables is $x$ and the neighborhood of $x$ in the constraint graph $G$ is $y, z, w, \ldots$, where $\varphi(y) \geq \varphi(z) \geq \varphi(w) \geq \ldots$

**five-vertex principle**. If (1) $\varphi(F)=3$ and $L_F(x)=2$, and (2) $\varphi(y)+\varphi(z)+\varphi(w)=5$, then

$$M(F) = M(F_1 \wedge x) \times M(F_2 \wedge x) + M(F_1 \wedge \neg x) \times M(F_2 \wedge \neg x) \quad (9)$$

where $F_1 = \{x, w\}$ and $F_2 = F/F_1$.

The aim of the principle is to remove $F_1$ such that $F$ doesn't contain the variables whose degree is 3. And since $F_1$ only contains two variables, it can be solved in polynomial time by exhaustive search. In fact, if $\varphi(x)=3$ and $L_F(x)=2$, then $\varphi(y)+\varphi(z)+\varphi(w) \geq 5$. This is because if $L_F(x)=2$, then $\varphi(y) \geq 2$ and $\varphi(z) \geq 2$. And if $\varphi(x)=3$ and $\varphi(y) \geq \varphi(z) \geq \varphi(w)$, then $\varphi(w)=1$. Therefore, when $\varphi(x)=3$ and $L_F(x)=2$, $\varphi(y)+\varphi(z)+\varphi(w) \geq 5$.

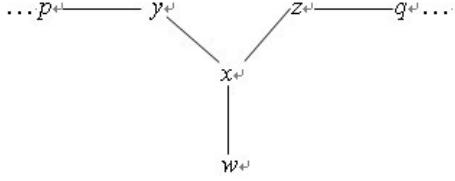

Figure 4: A constraint graph where $\varphi(x)=3$, $L_F(x)=2$ and $\varphi(y)+\varphi(z)+\varphi(w)=5$

## Algorithm $MC_2$ for Solving #2-SAT

The algorithm $MC_2$ for #2-SAT is based on the DPLL algorithm for satisfiablility modified to count all the satisfying assignments. The basic idea of the algorithm is to choose a variable and recursively count the number of satisfying assignments where the variable is *true* and the variable is *false*. We propose the framework of our algorithm $MC_2$ for #2-SAT in Figure 5. The algorithm employs a new principle to simplify formulae, i.e. the five-vertex principle. This allows us to eliminate variables whose degree is 3 in a formula, and therefore improve the efficiency of the algorithm. In addition, by transforming a formula into a constraint graph, we analyze the relationship between the adjacent variables in the constraint graph, which can choose better variables to branch. Note that in the algorithm $MC(F)$ is a function that solves the #2-SAT by exhaustive search. As we all know, if a #2-SAT is solved by exhaustive search, it will spend a lot of time. However, when the number of variables that the formula $F$ contains is so few, it may run in polynomial time. Therefore, we use the function $MC(F)$ only when the number of variables isn't above 4, which can guarantee the exhaustive search runs in polynomial time. In addition, since the operation on each node is the function $\Omega(F, R, l)$ running in polynomial time, we analyze the algorithm in Theorem 1 using the complexity measure described above.

**Theorem 1.** Algorithm $MC_2$ runs in $O(1.1892^m)$, where $m$ is the number of clauses.

Proof. Let us analyze the algorithm case by case.

Case 1, 2, and 3: These cases run in $O(1)$.

Case 4: This case doesn't increase the time needed.

Case 5.1: When $x$ is fixed a value, it splits $F$ into two paths and the clauses containing $x$ or $\neg x$ are removed. The worst case is only two clauses containing $x$ or $\neg x$. Since any two connected variables may form four clauses, we have $T(m) \leq 4T(\lceil m/2 \rceil -2)$, i.e. $T(m) \in O(m^{\log_2 4})=O(m^2)$.

---

Algorithm $MC_2(F, R)$

Case 1: $F$ has an empty clause. return 0.

Case 2: $F$ is empty. return $2^{|R|}$.

Case 3: $n \leq 4$. return $MC(F)$.

Case 4: $F$ consists of disjoint components $F_1$, $F_2$.
  return $MC_2(F_1, R) \times MC_2(F_2, R)$.

Case 5: $\varphi(F) \leq 2$.
  *1.* If $F$ is a path, choose $x$ to be a variable that can splits $F$ into paths of lengths $\lceil n/2 \rceil$ and $\lfloor n/2 \rfloor$.
  *2.* If $F$ is a cycle, choose $x$ arbitrary.
  return $MC_2(\Omega(F, R, \{x\}))+ MC_2(\Omega(F, R, \{\neg x\}))$.

Case 6: $\varphi(F)=3$ and $\varphi(x)=3$, where $N_F(x)=\{y, z, w\}$ and $\varphi(y) \geq \varphi(z) \geq \varphi(w)$
  *1.* If $L_F(x)=1$, return
    $MC_2(\Omega(F, R, \{U(x)\}))+ MC_2(\Omega(F, R, \{\neg U(x)\}))$.
  *2.* If $L_F(x)=2$ and $\varphi(y)+\varphi(z)+\varphi(w)=5$, return
    $MC_2(\Omega(F_1,R,\{x\})) \times MC_2(\Omega(F_2,R,\{x\}))+$
    $MC_2(\Omega(F_1, R, \{\neg x\})) \times MC_2(\Omega(F_2, R, \{\neg x\}))$,
    where $F_1=\{x, w\}$ and $F_2=F/F_1$.
  *3.* Otherwise, return
    $MC_2(\Omega(F, R, \{x\}))+MC_2(\Omega(F, R, \{\neg x\}))$.

Case 7: $\varphi(F) \geq 4$. Pick a variable $x$ such that $\varphi(x)=4$.
  return $MC_2(\Omega(F, R, \{x\}))+ MC_2(\Omega(F, R, \{\neg x\}))$.

Figure 5: $MC_2$ Algorithm

---

Case 5.2: When any variable is fixed a value, it can split $F$ into a path which case 5.1 is met. Therefore, we also have $T(m) \in O(m^2)$.

Case 6.1: Since $L_F(x)=1$, $\varphi(U(x)) \geq 2$. When $U(x)=true$, every clause containing $U(x)$ is removed and $\neg U(x)$ is removed from clauses. Then every clause containing $\neg U(x)$ can be also removed by function $\Omega$. Therefore, the current formula contains at least two clauses less than $F$. In addition, when $U(x)$ is fixed a value, $x$ forms a component containing three variables which meets the case 3. So we have $T(m)=2T(m-4)$ because the same situation is encountered when $U(x)=false$. This case takes $O(1.1892^m)$ time.

Case 6.2: Figure 4 describes this case. Since $F_1=\{x, w\}$, the number of satisfying assignments of $F_1$ can be counted in $O(1)$ by using $MC(F)$. In fact, the same process is carried out until $\varphi(F) \leq 2$, i.e. case 5 is met. Therefore, we have $T(m) \in O(m^2)$.

Case 6.3: In this case, $L_F(x) \geq 2$ and $\varphi(y)+\varphi(z)+\varphi(w) \geq 6$. The neighborhood of variable $x$ is $y$, $z$, and $w$. Since $L_F(x) \geq 2$, at least two of them are not just related to variable $x$. If we give a fix value to $x$, at least three clauses are removed. And simultaneously the clauses containing $y$ or $z$ or $w$ may be removed by function $\Omega$. Let $S=Unit(F, x) \cap \{y, z, w\}$ and $S'=Unit(F, \neg x) \cap \{y, z, w\}$. Then we have $T(m) = T(m-3-|S|) + T(m-3-|S'|)$. Since $\varphi(y)+\varphi(z)+\varphi(w) \geq 6$, $|S|+|S'| \geq 3$. Therefore, the worst case is when $T(m)=T(m-3)+T(m-6)$ with solution $O(1.1740^m)$.

Case 7: Since $\varphi(x)=4$, at least four clauses can be removed if $x$ is fixed a value. Therefore, we have $T(m)=2T(m-4)$ with solution $O(1.1892^m)$.

In total, $MC_2$ runs in $O(1.1892^m)$ time.

## Algorithm for #3-SAT

In this section, we present our algorithm $MC_3$ for solving #3-SAT and provide an upper bound $O(1.4142^m)$.

### Algorithm $MC_3$ for Solving #3-SAT

Algorithm $MC_3$ for #3-SAT is also based on the DPLL algorithm for satisfiablility modified to count all the satisfying assignments. We firstly present a notion used in this part. The frequency of a variable $x_i$ in a formula $F$ is the number of clauses in $F$ that $x_i$ appears in. Then we propose the framework of the algorithm $MC_3$ in Figure 6. The main idea of the algorithm is to choose the maximum frequency variable in all the 3-clauses to branch so that the input 3-clause formula is simplified into 2-clause formulae. Then we recursively count the number of satisfying assignments of these simplified 2-clauses by the algorithm $MC_2$. In the algorithm $MC_3$, there is a helpful function $\Omega(F, R, l)$ which has been described in the algorithm $MC_2$.

---

Algorithm $MC_3(F, R)$
Case 1: $F$ has an empty clause. return 0.
Case 2: $F$ is empty. return $2^{|R|}$.
Case 3: $F$ consists of disjoint components $F_1$, $F_2$.
return $MC_3(F_1, R) \times MC_3(F_2, R)$.
Case 4: If there exist 3-clauses in $F$, pick the maximum frequency variable $x$ in all the 3-clauses.
 return $MC_3(\Omega(F, R, \{x\})) + MC_3(\Omega(F, R, \{\neg x\}))$.
Case 5: Otherwise, return $MC_2(F, R)$.

---

Figure 6: $MC_3$ Algorithm

In fact, the algorithm $MC_3$ splits the search space into two parts. At first, it explores partial search tree until the 3-clause formula is transformed into 2-clause formulae. Then the algorithm $MC_2$ explores the complete search tree for the 2-clause formulae. Thus, the size of the search space of the $MC_3$ is equal to the DPLL-style algorithm which explores the complete search tree for an $n$-variable formula. Therefore, the algorithm $MC_3$ also can be solved in polynomial space. In addition, from the discussions above, we know that the algorithm $MC_2$ can be solved rapidly.

And the process of the transformation from 3-clause formula into 2-clause formulae is not difficult. As a result, the algorithm $MC_3$ improves the efficiency of solving the #3-SAT problem in a sense. In the next subsection, we will address the detailed complexity analysis about the algorithm $MC_3$.

### Complexity Analysis

In this subsection, we explain how to compute the complexity of the algorithm $MC_3$. As we have already described, we also use the branching tree to estimate the time complexity. However, the difference between the complexity analysis of the algorithm $MC_3$ and the others is that we only employ the branching tree to estimate time using in the process of the transformation from 3-clause formula into 2-clause formulae. When we acquire the time complexity of the simplified 3-clause formula, the time complexity of the algorithm $MC_3$ is easy to obtain by making use of the time complexity of the algorithm $MC_2$. The detailed proof will be presented in Theorem 2.

**Theorem 2.** Algorithm $MC_3$ runs in $O(1.4142^m)$, where $m$ is the number of clauses.

Proof. Let us analyze the algorithm in detail.
Case 1 and 2 can solve the problems completely. These cases run in $O(1)$.

Case 3 doesn't increase the time needed.

In Case 4, the maximum frequency of variables in 3-clauses is at least 2. Because if the maximum frequency of variables in 3-clauses is 1, it means that the frequencies of all the variables in 3-clauses are 1. Then the formula $F$ is mutually disjoint which case 3 is met. Thus, when $x$ is fixed a value, every clause containing $x$ (or $\neg x$) is either removed or simplified as 2-clauses. Since the maximal frequency of variables is at least 2, at least two clauses are removed when we give a fix value to $x$. Therefore, we have $T(m) = 2T(m-2)$ with solution $O(1.4142^m)$.

In Case 5, the formula only contains 2-clauses. We know that the algorithm $MC_2$ runs in $O(1.1892^m)$.

In total, the upper bound for the algorithm $MC_3$ is $O(1.4142^m)$.

## Conclusion

This paper addresses the worst-case upper bound for #2-SAT and #3-SAT problems with the number of clauses as the parameter. The algorithms presented are both DPLL-style algorithms. In order to improve the algorithms, we put forward a new five-vertex principle to simplify the formulae. After a skillful analysis of these algorithms, we obtain the worst-case upper bound $O(1.1892^m)$ for #2-SAT and $O(1.4142^m)$ for #3-SAT.

## Acknowledgement

This work was fully supported by the National Natural Science Foundation of China (Grant No. 60673099, 60803102, 60773097, and 60873146).